\documentclass{isprs} 
\usepackage{subfigure}
\usepackage{setspace}
\usepackage{geometry} 
\usepackage{epstopdf}
\usepackage[labelsep=period]{caption}  
\usepackage[british]{babel} 
\usepackage[hang]{footmisc}
\usepackage{makecell}
\usepackage{booktabs}
\usepackage{multirow}
\usepackage{url}
\usepackage{xcolor}
\usepackage{soul}
\sethlcolor{yellow}

\begin{document}

\title{Feasibility of Indoor Frame-Wise Lidar Semantic Segmentation via Distillation from Visual Foundation Model}
\date{}


\author{
Haiyang Wu\textsuperscript{1}, Juan J. Gonzales Torres\textsuperscript{1}, George Vosselman\textsuperscript{1}, Ville Lehtola\textsuperscript{1}
}

\address{
\textsuperscript{1 }Department of Earth Observation Science, Faculty of Geo-Information Science and Earth Observation (ITC),\\
University of Twente, 7522 NB Enschede, The Netherlands\\
haiyang.wu@utwente.nl; juznjogonzaleztorres@gmail.com; george.vosselman@utwente.nl; v.v.lehtola@utwente.nl
}
\abstract{
Frame-wise semantic segmentation of indoor lidar scans is a fundamental step toward higher-level 3D scene understanding and mapping applications. 
However, acquiring frame-wise ground truth for training deep learning models is costly and time-consuming. This challenge is largely addressed, for imagery, by Visual Foundation Models (VFMs) which segment image frames. The same VFMs may be used to train a lidar scan frame segmentation model via a 2D-to-3D distillation pipeline. The success of such distillation has been shown for autonomous driving scenes, but not yet for indoor scenes. Here, we study the feasibility of repeating this success for indoor scenes, in a frame-wise distillation manner by coupling each lidar scan with a VFM-processed camera image. The evaluation is done using indoor SLAM datasets, where pseudo-labels are used for downstream evaluation. Also, a small manually annotated lidar dataset is provided for validation, as there are no other lidar frame-wise indoor datasets with semantics.  Results show that the distilled model
achieves up to 56\% mIoU under pseudo-label evaluation and around 36\% mIoU with real-label, demonstrating the feasibility of
cross-modal distillation for indoor lidar semantic segmentation without manual annotations.
}


\maketitle


% ...
\section{Introduction}
Semantic information can significantly enhance indoor lidar-based environment modeling and understanding tasks~\cite{lehtola2022digital}. Representative applications include Scan-to-BIM (Building Information Modeling)~\cite{hu2024bim}, Scan-to-USD (Universal Scene Description)~\cite{halacheva2024scan2usd}, and semantic-metric SLAM~\cite{wu2025indoor,HybridVoxelMap2026}. However, due to the inherent sparsity and lack of texture in single-frame lidar point clouds, semantic information in these tasks is typically obtained either by performing semantic segmentation on scene-level point clouds~\cite{hu2020randla,wu2024point} or by leveraging additional visual sensor data~\cite{chen2025irs,soliman2025autonomous}. Scene-level segmentation is time-consuming and requires prior geometric modeling of the point cloud, which prevents it from contributing to the modeling process itself. Segmentation based on visual information relies on precise calibration to accurately project semantic masks onto point clouds, and only the overlapping field of view between the visual and lidar sensors can acquire semantic information, which is usually quite limited. 

In addition, some indoor semantic SLAM systems~\cite{wu2025indoor} perform frame-wise semantic segmentation using models pre-trained on complete indoor datasets such as S3DIS~\cite{armeni2016s3dis}, or infer semantics directly from geometric cues~\cite{bavle2023sgraph}. However, the former suffers from a significant domain gap between the pre-training and real-world indoor scenes, limiting its generalization, while the latter heavily depends on threshold-based heuristics and struggles to handle complex environments effectively.

Thanks to large-scale, high-quality outdoor autonomous driving datasets~\cite{caesar2020nuscenes,behley2019semantickitti} with frame-wise semantic annotations, many supervised models~\cite{puy2023waffle,wu2024point} have demonstrated strong performance in outdoor frame-wise lidar semantic segmentation task. However, such datasets are still lacking in indoor scenarios~\cite{bournez2024pushing}, and the substantial domain gap between indoor and outdoor environments makes it highly challenging to directly transfer supervised models trained on outdoor data to indoor applications.

Obtaining semantics without the need of having training data
is very attractive. The rapid development of 2D vision foundation models (VFMs)~\cite{jain2023oneformer,oquab2023dinov2,simeoni2025dinov3} has enabled the effective extraction of image features for various downstream tasks and has driven significant progress in cross-modal distillation from 2D to 3D under self-supervised learning paradigms~\cite{liu2023seal,puy2024scalr}. In this work, distillation refers to a learning process in which a student model learns from a teacher model by transferring feature knowledge instead of using labeled data. In cross-modal distillation, from 2D images to 3D point clouds, such feature knowledge is transferred between different modalities. In autonomous driving scenarios, this approach has achieved remarkable success in training 3D backbones, with the quality of the distilled 3D features approaching that of fully supervised methods~\cite{puy2024scalr}. Since cross-modal distillation does not rely on annotated datasets, it holds great potential for extension to indoor scenes to achieve effective semantic segmentation of single-frame lidar point clouds. 

However, this extension remains challenging: (1) unlike semantic segmentation in autonomous driving scenarios, which benefits from standardized large-scale benchmarks such as SemanticKITTI~\cite{behley2019semantickitti}, indoor scenes lack comparable datasets and established practices for single-frame semantic segmentation; and (2) these frameworks still require partial data for downstream supervision, yet frame-wise annotated indoor lidar datasets are scarce.

\sloppy

Frame-wise segmentation enables direct per-frame semantic understanding and supports online processing, unlike scene-level indoor segmentation, which requires global reconstruction and is unsuitable for real-time tasks such as semantic SLAM. To this end, we employ an advanced cross-modal distillation framework ScaLR~\cite{puy2024scalr}, originally developed for autonomous driving scenarios, and study its adaptation and feasibility in indoor scenes. Specifically, our main contributions are summarized as follows:

\begin{enumerate}
    \item We develop a dedicated data processing pipeline based on indoor SLAM datasets to enable the adaptation of ScaLR in indoor settings.
    \item We generate pseudo-labels via projections from VFM to supervise downstream tasks, demonstrating the feasibility of training and fine-tuning on arbitrary datasets without manual annotations. Additionally, we manually annotate a small indoor dataset to validate the reliability of pseudo-label evaluation and assess the model’s real performance.
\end{enumerate}
\sloppy

The paper is organized as follows. We review the related work in the Section \ref{sec:related_work}. In Section \ref{sec:framework_overview}, we describe the ScaLR~\cite{puy2024scalr} framework that we apply to four unlabeled indoor multi-sensor SLAM datasets TIERS, NTU-VIRAL, M2DGR, and ITC, described in Section \ref{sec:experiments}. The purpose is to (1) distill feature knowledge and then (2) to learn pseudo-labels from VFM using linear probing and fine tuning. The experimental results and validation against a small set of manually annotated ground truth are also presented in Section \ref{sec:experiments}. Section \ref{sec:discussion} discusses the findings and limitations, and Section \ref{sec:conclusions} concludes the paper.

\section{Related work}
\label{sec:related_work}
We categorize the related work in supervised and distilled methods. When reading this section, the reader should note that point cloud segmentation methods can in principle be used to segment both large, registered point clouds and lidar scan frames, but if a model is trained on one, it does not perform on the other because of the domain gap in between.

\subsection{Supervised Lidar Point Cloud Semantic Segmentation}

Since the introduction of Point Transformer architectures~\cite{zhao2021pt}, direct point-based processing has become an effective paradigm for lidar point cloud segmentation. The latest Point Transformer V3~\cite{wu2024point} simplifies the transformer design through point serialization and patch attention, achieving state-of-the-art (SOTA) results on large-scale lidar benchmarks. In contrast, voxel-based methods such as MinkUNet~\cite{choy20194d} and NUC-Net~\cite{wang2025nuc} discretize the 3D space to enable efficient sparse convolution, providing a good trade-off between accuracy and computational cost. Meanwhile, projection-based approaches like WaffleIron~\cite{puy2023waffle} project point clouds onto regular 2D grids and perform dense convolution for fast inference while retaining competitive segmentation accuracy.

Despite their success, all these supervised lidar segmentation frameworks rely heavily on dense human annotations, motivating recent progress in cross-modal and self-supervised learning, including distillation.

\subsection{Cross-Modal Distillation for Lidar Point Cloud Semantic Segmentation}

The rapid progress of VFMs~\cite{jain2023oneformer,simeoni2025dinov3} has provided powerful feature representations that can be transferred to 3D scene understanding tasks. Building on these advances, recent research has focused on cross-modal distillation for lidar segmentation in autonomous driving scenarios, where features from pretrained 2D VFMs is distilled into 3D lidar backbones to reduce manual ground truth annotation work.

Early work such as PPKT~\cite{liu2021ppkt} introduced the point–pixel contrastive framework, transferring semantic representations from 2D image features to 3D point features through pixel-to-point correspondences. Extending this idea, SLiDR~\cite{sautier2022slidr} proposed structured distillation using superpixels and superpoints with a contrastive loss, marking the first large-scale application of this paradigm to autonomous driving lidar data. SEAL~\cite{liu2023seal} further enhanced the framework by enforcing temporal consistency across sequential point clouds, leveraging superpoint clustering and motion-aware contrastive losses to improve representation stability. Building upon these foundations, M3Net~\cite{liu2024m3net} introduced multi-space alignment, including feature, data, and label-space alignment, supported by a composite loss design that improved generalization across datasets. ScaLR~\cite{puy2024scalr} revisits vision-to-lidar distillation by emphasizing three key factors: model capacity, teacher quality, and dataset diversity, and adopts a simple cosine-similarity loss with random camera sampling. By scaling both 2D and 3D networks and training across multiple lidar datasets, it achieves robust, label-efficient representations that approach fully supervised performance. SuperFlow++~\cite{xu2025superflow++} extends this line of research by incorporating spatiotemporal consistency into cross-modal pretraining, aligning lidar and camera features across consecutive frames.

The above cross-modal distillation frameworks have achieved remarkable success in autonomous driving scenarios, where lidar data are processed in a frame-wise manner. In contrast, similar paradigms are now being extended to indoor scene understanding, where datasets such as S3DIS~\cite{armeni2016s3dis} provide fully reconstructed scene-level point clouds rather than frame-wise lidar scans. Building upon this setting, PointDC~\cite{chen2023pointdc} transfers multi-view features from self-supervised 2D models to indoor scene point clouds through cross-modal distillation and iterative super-voxel clustering, enabling unsupervised segmentation on scene-level datasets such as ScanNet-v2~\cite{dai2017scannet} and S3DIS~\cite{armeni2016s3dis}. Extending this paradigm, D-DITR~\cite{zeid2025dino} unifies 2D-to-3D feature injection and distillation within a single framework, leveraging frozen DINOv2~\cite{oquab2023dinov2} representations to enhance both indoor and outdoor 3D segmentation without labeled data.

Cross-modal distillation has proven effective in both indoor and outdoor 3D scene understanding. However, due to the lack of high-quality frame-wise lidar semantic segmentation benchmarks and the large domain gap between indoor and outdoor data, as well as between scene-level and frame-wise indoor point clouds, its feasibility for indoor frame-wise lidar segmentation remains unexplored. Some related studies improve cross-domain lidar segmentation using pseudo-labels~\cite{kaushik2025unsupervised}, but they are limited to outdoor datasets. Building upon these advances, we investigate the feasibility of applying cross-modal distillation to indoor lidar semantic segmentation at the frame level, aiming to enable real-time and per-frame semantic understanding. 

\begin{figure}[t]
\centering
\includegraphics[width=\linewidth]{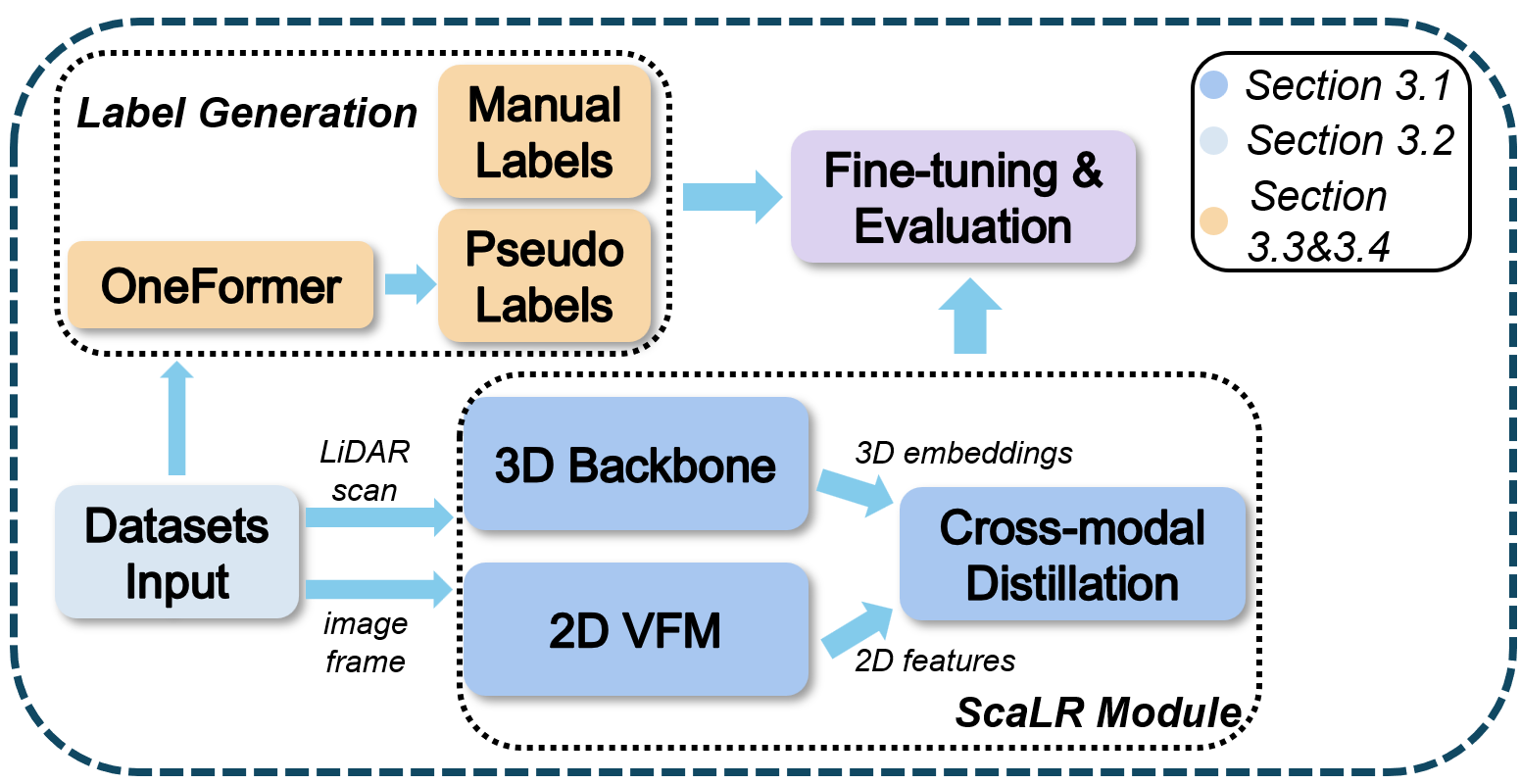}
\caption{Workflow of ScaLR adaptation to indoor datasets.}
\label{fig:overview}
\end{figure}

\section{Framework Overview}
\label{sec:framework_overview}
Figure~\ref{fig:overview} illustrates the overall workflow of the proposed framework. It follows the structure of the ScaLR framework~\cite{puy2024scalr}, originally developed for outdoor autonomous driving scenes, but here it is adapted to indoor SLAM datasets to evaluate its feasibility for frame-wise indoor segmentation tasks. The framework consists of three main components: cross-modal distillation, indoor data adaptation, and label generation.
Each component is described in the following subsections.

\subsection{Cross-modal Distillation}
The core idea of cross-modal distillation is to transfer features from a pretrained VFM in the 2D domain to a 3D network for point cloud understanding. 
In this work, we directly employ the existing ScaLR framework without altering its core mechanisms. The framework establishes a \textit{teacher–student} relationship, where the VFM (DINOv2~\cite{oquab2023dinov2}) acts as the teacher providing meaningful pixel-level descriptors, and the 3D model (WaffleIron~\cite{puy2023waffle}) serves as the student that learns point-wise semantic representations from them.

Given an aligned pair of image and lidar point cloud, each 3D point $p_i$ is projected to its corresponding image pixel $\rho(i)$. The feature vectors $\tilde{f}_i$ and $g_{\rho(i)}$ are extracted respectively from the 3D student and 2D teacher networks. To align their feature representations, the learning objective minimizes the distance between normalized descriptors:
\begin{equation}
\mathcal{L}_{\mathrm{sim}} = \frac{1}{N}\sum_{i=1}^{N} \left\| \tilde{f}_i - g_{\rho(i)} \right\|_2,
\label{eq:loss}
\end{equation}
where both $\tilde{f}_i$ and $g_{\rho(i)}$ are $\ell_2$-normalized, and $N$ denotes the total number of valid points in the batch. 
This unsupervised loss enforces the point features to be semantically consistent with their corresponding image pixels, effectively embedding prior semantic context into the 3D representation.

In practice, image inputs are first resized (e.g., $448\times224$) before being passed through the 2D teacher model, and then bilinearly interpolated back to the original resolution to ensure matching with the lidar projection. This design avoids the need for superpixels or superpoints and has shown competitive performance against earlier distillation methods such as PPKT~\cite{liu2021ppkt} and SLiDR~\cite{sautier2022slidr}. By leveraging pretrained VFMs like DINOv2~\cite{oquab2023dinov2} or OneFormer~\cite{jain2023oneformer}, the ScaLR framework enables a self-supervised pretraining of the 3D backbone without requiring any manually labeled 3D data. 

\subsection{Indoor Data Adaptation}
To enable the ScaLR framework to operate on indoor SLAM datasets, a dedicated data adaptation pipeline was developed. This pipeline focuses on structuring and importing data from indoor SLAM datasets while preserving the original distillation workflow. Specifically, the implementation supports datasets such as NTUviral~\cite{nguyen2022ntu}, TIERS~\cite{qingqing2022tiers}, and M2DGR~\cite{yin2021m2dgr}, which provide synchronized lidar–camera pairs suitable for cross-modal learning. 

\begin{figure}[htbp]
\centering
\includegraphics[width=\linewidth, height=0.15\textheight]{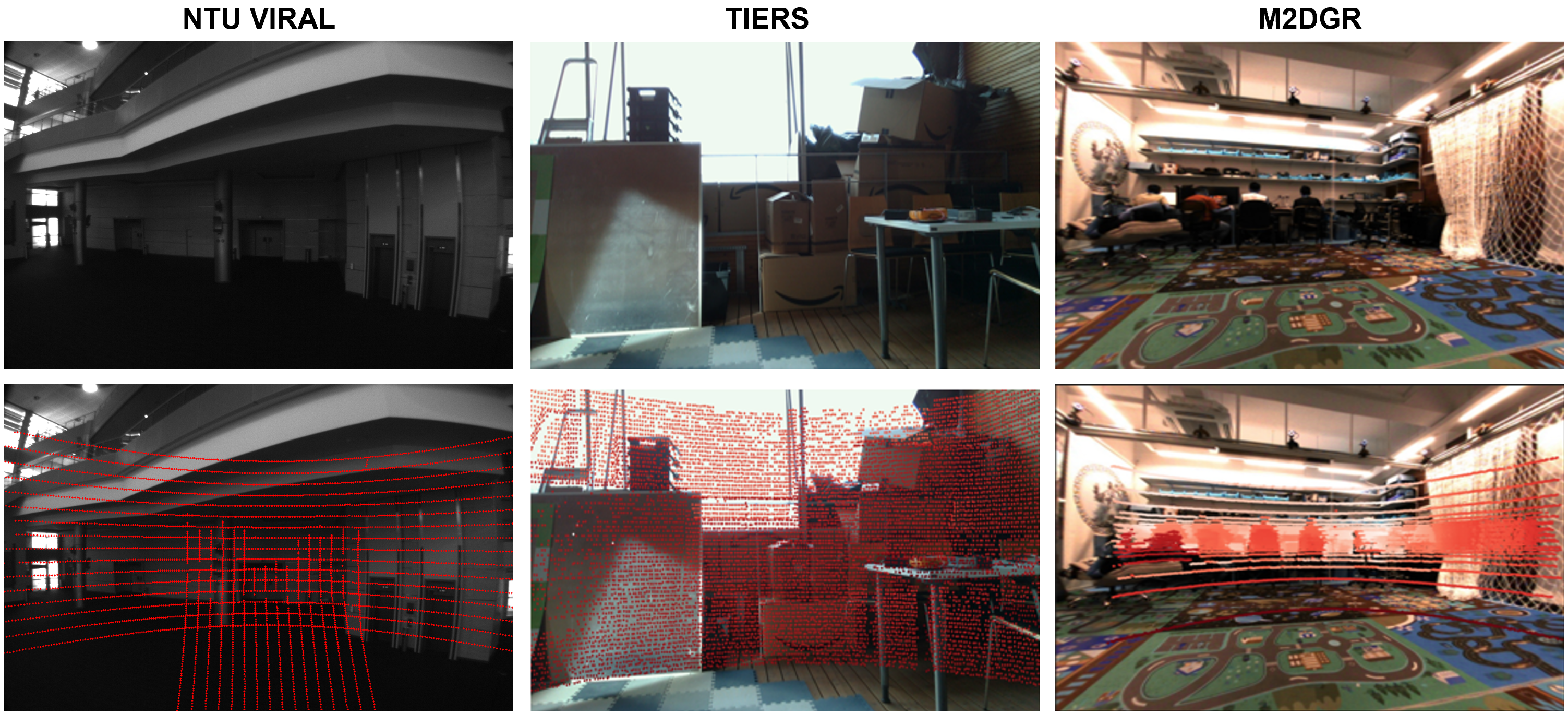}
\caption{Comparison of lidar scan projections in different indoor SLAM datasets.}
\label{fig:projection-comparison}
\end{figure}

Each dataset exhibits unique sensing configurations and lidar scan patterns, resulting in distinct projection geometries when aligning point clouds with images. For instance, NTUviral employs a dual-lidar setup with complementary vertical and horizontal orientations, while TIERS and M2DGR use single lidar sensors with different angular resolutions and ranges. An illustration of these variations and the corresponding projection alignment is provided in Figure~\ref{fig:projection-comparison}. 

The framework was extended with new data-loading modules to ensure compatibility with these heterogeneous sensor setups. Each dataset is managed through two dedicated modules responsible for data preparation in the distillation and downstream task stages. These modules follow the same design as the generic ScaLR dataset interface, enabling consistent handling of lidar–image pairs and class mappings across datasets. 

During preprocessing, each lidar scan is associated with its temporally closest image using dataset-specific timestamps. 
The projection from lidar coordinates to the image plane is performed through intrinsic and extrinsic transformations defined for each dataset, ensuring that every 3D point $p_i$ is linked to its corresponding pixel $\rho(i)$. 
This projection enables the distillation loss in Eq.~(\ref{eq:loss}) to be computed consistently across all datasets. 

To standardize the training configuration, all datasets are reformatted into a unified directory and annotation structure compatible with ScaLR. This allows flexible composition of training and evaluation subsets, including the possibility of merging multiple datasets for distillation and fine-tuning.

\subsection{Pseudo-label Generation}
To compensate for the lack of labeled indoor frame-wise lidar data, pseudo-labels are generated automatically by projecting the 2D semantic segmentation outputs from VFM onto the corresponding lidar point clouds. For each synchronized image–point cloud pair, the segmentation mask produced by the OneFormer~\cite{jain2023oneformer} assigns a semantic label to each pixel, which is then transferred to the nearest projected 3D point. 

\begin{figure}[htbp]
\centering
\includegraphics[width=\linewidth, height=0.10\textheight]{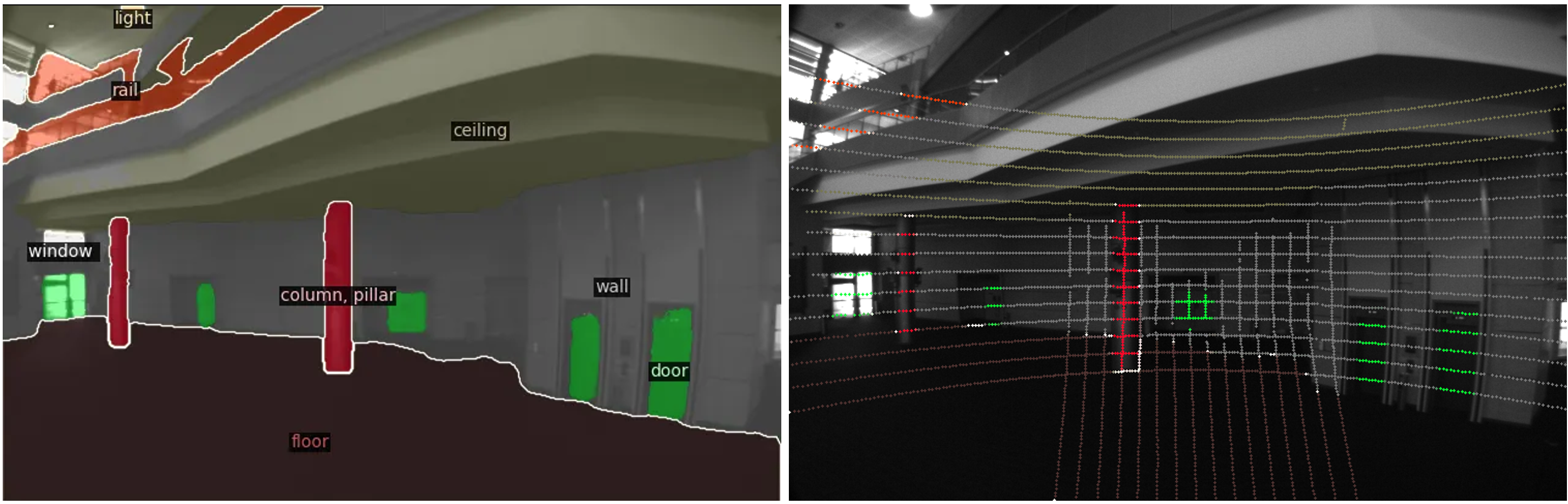}
\caption{Example of cross-modal alignment between OneFormer segmentation (left) and lidar-projected semantics (right) on the NTU-VIRAL dataset. The lidar points are colorized according to the corresponding segmentation mask.}
\label{fig:pseudo_labels}
\end{figure}

This approach allows the framework to utilize large-scale indoor SLAM datasets~\cite{nguyen2022ntu,yin2021m2dgr,qingqing2022tiers} without manual annotation. In this study, the resulting pseudo-labels serve two purposes: they provide supervision for downstream linear probing and fine-tuning, and they also serve as a proxy reference for evaluation on datasets without manual frame-wise labels. An example of the pseudo-labeled projection results is shown in Figure~\ref{fig:pseudo_labels}.

\subsection{Real-label Generation}
To validate the reliability of pseudo-labels and quantitatively assess segmentation accuracy, a small portion of real-labeled data was manually created. Specifically, 1,720 lidar frames (approximately 15\% of the total data) from three sequences of the ITC indoor dataset, collected in a university building environment, were manually annotated using the open-source software CloudCompare\footnote{\url{http://www.cloudcompare.org}}, producing accurate per-point semantic labels for the testing subset. 
\begin{figure}[htbp]
\centering
\includegraphics[width=\linewidth]{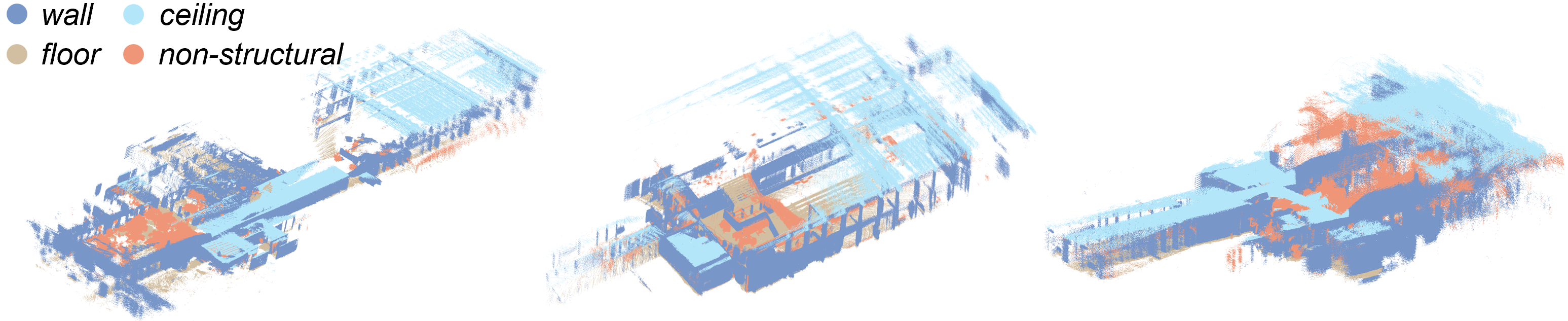}
\caption{Manually labeled results on the ITC dataset.}
\label{fig:manual-label-results}
\end{figure}

These ground-truth annotations are used solely for evaluation, ensuring an objective comparison between pseudo-labeled and manually labeled results. Although this manually labeled subset provides an important real-label validation, it is limited to 1,720 frames from three sequences collected in a single building. An illustration of the manually labeled point clouds is shown in Figure~\ref{fig:manual-label-results}.

\section{Experiments and Results}
\label{sec:experiments}
\subsection{Data}
\label{sec:experimental-setup}
We evaluate on three public indoor datasets: NTU-VIRAL~\cite{nguyen2022ntu}, TIERS~\cite{qingqing2022tiers}, and M2DGR~\cite{yin2021m2dgr}, as well as the ITC dataset collected in an university building environment using a handheld lidar device. The selected sequences and lidar configurations are summarized in Table~\ref{tab:datasets}. For all datasets, the data are split approximately into 70\% training, 15\% validation, and 15\% testing subsets. Each lidar frame is paired with its corresponding RGB image, enabling 2D-to-3D distillation for frame-wise segmentation.

\begin{table}[htbp]
  \centering
  \vspace{-0.5em}
  \begin{tabular}{l l l c}
    \toprule
    \textbf{Dataset} & \textbf{Sequences} & \textbf{lidar sensor} & \textbf{Scans} \\
    \midrule
    NTU-VIRAL & \textit{eee}, \textit{nya}, \textit{sbs} & Ouster OS1-16 & 21000 \\
    TIERS & all & Ouster OS1-16 & 8000 \\
    M2DGR & indoor & Velodyne VLP-32 & 27500 \\
    ITC & all & Hesai XT-32 & 11533 \\
    \bottomrule
  \end{tabular}
  \caption{Experimental datasets and lidar configurations.}
  \label{tab:datasets}
\end{table}

\subsection{Distillation details}
The training runs for 25~epochs with a learning rate of~0.002, weight decay of~0.03, and batch size of~8. For linear probing, only the classifier is trained using a learning rate of~0.001, weight decay of~0.003, and 20~epochs. For fine-tuning, the entire model is optimized with a learning rate of~0.002, layer decay~0.99.

All experiments were conducted on a server node equipped with two NVIDIA~A40 GPUs (46\,GB each). 


\subsection{Linear Probing}
Pseudo-labels generated from OneFormer model~\cite{jain2023oneformer} provide supervision for both linear probing and fine-tuning stages. We first evaluate the cross-modal representations using linear probing, following the ScaLR workflow. 
Note that after distilling the feature knowledge, which only helps in feature extraction, the linear probing or fine tuning is required to improve the the model in connecting these distilled features into the correct semantic segmentation output.

The results in Table~\ref{tab:cross-distillation} show the mean IoU on four downstream datasets when the model is distilled on different sources.

The results reveal a clear domain dependence: the best performance consistently lies along the diagonal, indicating that representations transfer poorly across datasets even within indoor domains. Nevertheless, pre-training on the combined All Indoor set substantially improves generalization, achieving 39.38\% on NTU-VIRAL, 48.08\% on TIERS, and 34.8\% on M2DGR. This highlights the benefit of jointly leveraging heterogeneous indoor data, suggesting a step toward a more universal lidar representation. Meanwhile, the considerable performance drop on outdoor data (NuScenes) underscores a persistent domain gap between indoor and outdoor environments.
\begin{table}[htbp]
\centering
\setlength{\tabcolsep}{4pt}
\begin{tabular}{lccc|c}
\toprule
 & \multicolumn{4}{c}{\textbf{Downstream Dataset}} \\
\cmidrule(lr){2-5}
\textbf{\makecell{Pretraining\\Dataset}} & \textbf{NTUviral} & \textbf{TIERS} & \textbf{M2DGR} & \textbf{Outdoor$^*$} \\
\midrule
NTU-VIRAL   & \textbf{40.71\%} & 34.12\% & 25.84\% & 28.23\% \\
TIERS      & 33.34\% & \textbf{49.68\%} & 27.24\% & 32.48\% \\
M2DGR      & 33.64\% & 34.71\% & \underline{33.36\%} & 32.18\% \\
All Indoor & \underline{39.38\%} & \underline{48.08\%} & \textbf{34.8\%} & 29.21\% \\
\bottomrule
\end{tabular}
\caption{Mean IoU (\%) for linear probing on various downstream datasets under different pre-training datasets. “All Indoor” includes NTU-VIRAL, TIERS, and M2DGR datasets. $^*$Outdoor data is from NuScenes.}
\label{tab:cross-distillation}
\vspace{2mm}
\end{table}

\subsection{Fine-tuning}
We further evaluate the adaptability of the distilled representations through fine-tuning on each indoor dataset using pseudo-label supervision. Table~\ref{tab:direct-finetune} summarizes the resulting performance. NTU-VIRAL and TIERS achieve comparable performance, reaching about 51--52\% mIoU and over 85\% overall accuracy. 
The M2DGR dataset shows lower mIoU (39.8\%), likely due to its limited scene diversity and smaller spatial coverage. Overall, these results demonstrate that fine-tuning substantially enhances segmentation performance and partially mitigates the domain gap observed in the linear probing stage.

\begin{table}[htbp]
\centering
\begin{tabular}{lccc}
\toprule
\textbf{Dataset} & \textbf{mIoU (\%)} & \textbf{mAcc (\%)} & \textbf{oAcc (\%)} \\
\midrule
NTU-VIRAL & 51.4 & 63.0 & 86.0 \\
TIERS    & 51.1 & 62.5 & 85.2 \\
M2DGR    & 39.8 & 48.2 & 85.3 \\
\bottomrule
\end{tabular}
\caption{Quantitative results (mIoU, mAcc, oAcc) of fine-tuned models on the test sets.}
\label{tab:direct-finetune}
\end{table}

\subsection{Real-Label Evaluation on ITC Dataset}
\label{sec:itc-evaluation}
To assess the reliability of pseudo-label evaluation, we perform an in-domain validation on the ITC dataset, which comprises diverse indoor scenes collected in a university building. The ScaLR model is both distilled and fine-tuned on the ITC data using pseudo-label supervision under a four-class structural mapping with \textit{wall}, \textit{floor}, \textit{ceiling}, and \textit{non-structural}. The pseudo labels are obtained by projecting the OneFormer image predictions onto the lidar points and then remapping the resulting categories into this four-class label space. In the pseudo-label mapping, \textit{wall} includes \textit{wall} and \textit{building}, \textit{floor} includes \textit{floor}, \textit{sidewalk}, and \textit{road}, and some categories such as chair, desk, and furniture are ignored rather than assigned to \textit{non-structural}. In the manually annotated ITC labels, by contrast, all categories other than \textit{wall}, \textit{floor}, and \textit{ceiling} are grouped into \textit{non-structural}. Its segmentation outputs on the ITC test set are evaluated against pseudo labels derived from OneFormer projections and manually annotated real labels. Since no public benchmark exists for indoor frame-wise lidar semantic segmentation, RandLA-Net~\cite{hu2020randla} is included only as a cross-domain supervised reference.

Table~\ref{tab:scalr-evaluation} summarizes the evaluation results. Against pseudo labels, the ScaLR model achieves 56.5\% mIoU, 70.9\% mAcc, and 77.6\% overall accuracy. Against the manually annotated real labels, the same model yields 35.8\% mIoU and 70.3\% overall accuracy. This gap is influenced not only by pseudo-label noise, likely caused by imperfect 2D segmentation, projection misalignment, and inconsistencies near object boundaries or sparsely sampled regions, but also by the mismatch between the pseudo-label and real-label mappings, especially in the definition of \textit{non-structural}.

RandLA-Net, pre-trained on S3DIS~\cite{armeni2016s3dis} without exposure to ITC data, achieves only 10.7\% mIoU on the same test set. Per-class results in Table~\ref{tab:scalr-perclass} further show that ScaLR captures dominant structural categories such as \textit{walls} (67.9\%) and \textit{floors} (28.4\%) more effectively than the supervised baseline.

Overall, these findings confirm that pseudo-label-based evaluation provides a useful and consistent proxy for assessing semantic segmentation performance in the absence of dense ground-truth annotations. Moreover, they demonstrate that cross-modal distillation effectively transfers feature representations into the 3D domain, enabling robust and transferable indoor lidar segmentation.
\begin{table}[htbp]
\centering
\setlength{\tabcolsep}{3pt}
\begin{tabular}{lcccc}
\toprule
\textbf{Method} & \textbf{Ref.} & \textbf{mIoU (\%)} & \textbf{mAcc (\%)} & \textbf{oAcc (\%)} \\
\midrule
ScaLR & Pseudo & 56.5 & 70.9 & 77.6 \\
ScaLR & Real & 35.8 & 46.3 & 70.3 \\
RandLA-Net & Real & 10.7 & 21.4 & 26.8 \\
\bottomrule
\end{tabular}
\caption{Performance comparison for point cloud semantic segmentation. Reference label (Ref.) is either pseudo or real label.}
\label{tab:scalr-evaluation}
\vspace{2mm}
\end{table}

\begin{table}[htbp]
\centering
\setlength{\tabcolsep}{3pt}
\begin{tabular}{lcccc}
\toprule
\textbf{Method} & \textbf{Wall} & \textbf{Floor} & \textbf{Ceiling} & \textbf{Non-structural} \\
\midrule
ScaLR & 67.9 & 28.4 & 10.2 & 36.5 \\
RandLA-Net & 17.8 & 0.0 & 1.2 & 23.8 \\
\bottomrule
\end{tabular}
\caption{Per-class IoU (\%) on the ITC dataset based on real-label evaluation.}
\label{tab:scalr-perclass}
\end{table}

\subsection{Sensitivity to the VFM}
We investigate the influence of the VFM used for distillation by comparing OneFormer~\cite{jain2023oneformer} 
and DINOv2~\cite{oquab2023dinov2} on the ITC and TIERS datasets. In this comparison, all fine-tuning stages rely on pseudo labels generated from OneFormer projections to ensure consistent supervision. The goal is not to rank VFMs exhaustively, but to examine whether the distilled 3D representation remains stable under different 2D teachers.

Results in Table~\ref{tab:foundation-itc-tiers} show that both VFMs lead to comparable segmentation accuracy. On the ITC dataset, DINOv2 slightly outperforms OneFormer by 1.5\,mIoU and 1.9\,mAcc, while on TIERS the two models yield nearly identical performance with marginal differences below 0.5\%. This indicates that the ScaLR framework remains stable across different VFMs, and that both vision transformer–based and segmentation-oriented encoders can provide effective supervision for indoor lidar segmentation.
\begin{table}[htbp]
\setlength{\tabcolsep}{3pt}
\centering
\begin{tabular}{llccc}
\toprule
\textbf{Dataset} & \textbf{Foundation} & \textbf{mIoU (\%)} & \textbf{mAcc (\%)} & \textbf{oAcc (\%)} \\
\midrule
\multirow{2}{*}{ITC} 
& OneFormer & 55.0 & 69.0 & 76.8 \\
& DINOv2     & 56.5 & 70.9 & 77.6 \\
\midrule
\multirow{2}{*}{TIERS} 
& OneFormer & 67.8 & 77.1 & 85.4 \\
& DINOv2     & 67.8 & 77.6 & 85.0 \\
\bottomrule
\end{tabular}
\caption{Performance comparison (mIoU, mAcc, oAcc) of models distilled with different VFMs on the ITC and TIERS datasets. 
All results are obtained after fine-tuning with pseudo labels generated from OneFormer projections.}
\label{tab:foundation-itc-tiers}
\end{table}

\begin{figure}[htbp]
\centering
\includegraphics[width=\linewidth]{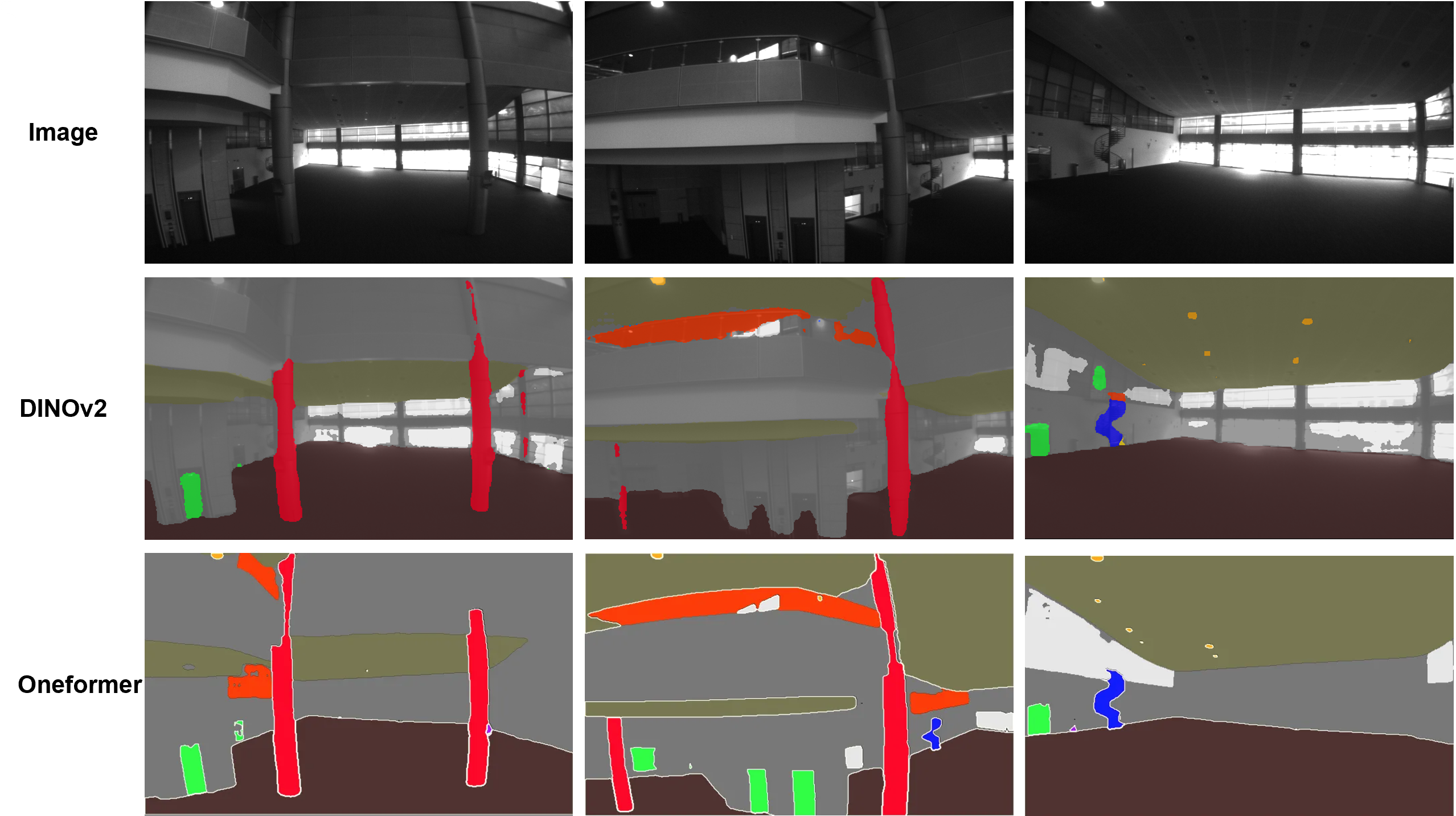}
\caption{Qualitative comparison of segmentation outputs from DINOv2 (with Mask2Former head) and OneFormer on the NTU-VIRAL dataset.}
\label{fig:segmentation-comparison}
\end{figure}

Figure~\ref{fig:segmentation-comparison} provides a qualitative comparison between the semantic segmentation outputs of DINOv2 (with a Mask2Former~\cite{cheng2022masked} head) and OneFormer on the NTU-VIRAL dataset. The visual consistency observed across structural categories supports the quantitative findings that both VFMs offer stable supervision for indoor lidar segmentation, with only minor variations in fine-scale details.

\subsection{Training Efficiency and Runtime Performance}
We further analyze the computational efficiency of the 3D backbone by comparing WaffleIron~\cite{puy2023waffle} models with different depths. 
Table~\ref{tab:waffleiron-depth} summarizes the fine-tuning time, GPU memory consumption, segmentation accuracy, 
and inference speed across configurations of 24, 36, and 48 layers.

The results reveal that deeper networks bring only marginal accuracy improvements while substantially increasing both training cost and memory usage. Specifically, the 48-layer model achieves the highest mIoU of 51.8\% but requires 8.4\,hours and 76\,GB of memory, compared with 3.4\,hours and 50\,GB for the 24-layer version. 
Inference speed drops from 15.4\,Hz to 9.8\,Hz as depth increases. Considering the small accuracy gain, the 24-layer configuration provides the best trade-off between performance and efficiency 
for practical indoor segmentation.
\begin{table}[htbp]
\centering
\setlength{\tabcolsep}{4pt}
\begin{tabular}{lcccc}
\toprule
\textbf{Depth} & \textbf{FT (h)} & \textbf{Mem. (GB)} & \textbf{mIoU (\%)} & \textbf{Inf. spd. (Hz)} \\
\midrule
24 & 3.41 & 50.1 & 50.9 & 15.41 \\
36 & 5.73 & 66.3 & 51.0 & 12.17 \\
48 & 8.39 & 76.0 & 51.8 & 9.84 \\
\bottomrule
\end{tabular}
\caption{Comparison of computational efficiency and resource usage among WaffleIron models of different depths.}
\label{tab:waffleiron-depth}
\end{table}

\section{Discussion}
\label{sec:discussion}
\subsection{Feasibility and Key Findings}

As summarized in Table~\ref{tab:direct-finetune}, the distilled model attains around 50\%~mIoU and over 85\%~overall accuracy on NTU-VIRAL and TIERS. Since, to the best of our knowledge, frame-wise indoor lidar semantic segmentation has not been previously studied and no labeled datasets or established benchmarks exist for this task, direct quantitative comparison is not possible. These results therefore serve as an initial reference, demonstrating that the framework yields stable and transferable semantic representations even without manual labels. The results on the ITC dataset further validate this feasibility: when evaluated against real annotations, the fine-tuned model reaches 35.8\%~mIoU and 70.3\%~overall accuracy, outperforming the fully supervised RandLA-Net baseline by a large margin (10.7\%~mIoU). Although the absolute scores are lower than those typically achieved on outdoor benchmarks, they confirm that the distilled representations capture meaningful indoor semantics and that cross-modal distillation provides a feasible starting point for label-free indoor lidar semantic segmentation.

The framework remains robust across different visual foundation models. As shown in Table~\ref{tab:foundation-itc-tiers}, both OneFormer and DINOv2 yield comparable segmentation accuracy, indicating that the learned 3D features are not sensitive to the specific 2D encoder. The efficiency analysis in Table~\ref{tab:waffleiron-depth} also highlights the practicality of the approach: the 24-layer WaffleIron backbone provides the best balance between accuracy and computational cost, achieving real-time inference at over 15~Hz.


\subsection{Domain Generalization and Transferability}
The cross-dataset evaluations reveal that the distilled representations remain strongly domain-dependent. As shown in Table~\ref{tab:cross-distillation}, the best performance consistently appears along the diagonal, while cross-domain results degrade notably when the model is evaluated on unseen indoor datasets. For instance, a model distilled on NTU-VIRAL achieves 40.7\%~mIoU on its native domain but only around 25–33\% on other datasets. This confirms that geometric and appearance variations across indoor scenes introduce a substantial domain gap.

The observed domain sensitivity mainly arises from the diversity of sensor configurations and scene characteristics among the indoor datasets. Differences in lidar hardware, such as the number of multi-line channels and field of view, directly affect point density and geometric resolution. In particular, lidars with fewer lines produce sparser vertical sampling, while lidars with more lines provide more complete geometric coverage. This affects the point distribution, especially near object boundaries and thin structures, and reduces transferability across datasets~\cite{chen2025m3net}. For example, NTU-VIRAL employs a dual-lidar setup paired with a monochrome camera, while TIERS and M2DGR use single multi-channel lidar configurations with an RGB camera. These disparities lead to distinct spatial sampling patterns and geometric distributions that reduce cross-domain feature compatibility. Scene-specific factors, including building layout and object composition, further accentuate this variation, resulting in limited feature transferability across datasets.

Training on the combined All Indoor set alleviates this issue to some extent. As summarized in Table~\ref{tab:cross-distillation}, pre-training on heterogeneous indoor data improves the mean IoU across all target domains, reaching 39.4\% on NTU-VIRAL, 48.1\% on TIERS, and 34.8\% on M2DGR. This demonstrates that combining complementary datasets helps the model learn more transferable 3D feature representations while maintaining strong in-domain performance. These results indicate that incorporating diverse sensor modalities and scene types during distillation is an effective strategy to enhance domain robustness across indoor scenes.

\subsection{Reliability of Pseudo-label Evaluation}
The reliability of pseudo-label-based evaluation is examined on the ITC dataset by comparing model performance under pseudo and manually annotated labels. As shown in Table~\ref{tab:scalr-evaluation}, the model achieves 56.5\%~mIoU when evaluated against pseudo-labels but only 35.8\%~mIoU with manual annotations, while the overall accuracy remains comparable. This noticeable gap indicates that pseudo-labels tend to overestimate model performance due to label noise and incomplete category consistency. The discrepancy likely comes from imperfect 2D teacher predictions, projection misalignment between pixels and points, and semantic ambiguity near thin structures, small objects, and class boundaries. Nevertheless, the general performance tendency remains comparable to that under real-label evaluation, suggesting that pseudo-label assessment can approximate true model behavior within reasonable bounds when dense human annotations are unavailable. In future work, simple refinement strategies such as confidence-based filtering or temporal consistency checks may help improve pseudo-label quality.

\subsection{Limitations and Future Work}
Although this study demonstrates the feasibility of applying cross-modal distillation to indoor frame-wise lidar segmentation, several limitations remain:
\begin{itemize}
    \item \textbf{Dataset diversity.} The experiments rely on a limited number of indoor datasets, each with distinct sensor configurations and scene characteristics, which restricts large-scale generalization and cross-domain consistency. In particular, the manually labeled ITC subset is small and collected in a single building, limiting the diversity of real-label evaluation.
    \item \textbf{Pseudo-label accuracy.} The pseudo-labels used for downstream tasks contain noticeable noise and class imbalance, resulting in substantial performance discrepancies when compared with manual annotations. This noise likely stems from imperfect 2D segmentation, projection misalignment, and semantic ambiguity near sparse or boundary regions.
    \item \textbf{Temporal independence.} The current framework processes lidar scans in a purely frame-wise manner without leveraging a temporal {\it sliding window} or multi-view information, which may limit geometric continuity across consecutive frames. This is particularly relevant for indoor SLAM sequences, where adjacent frames usually have substantial overlap and simple temporal smoothing or fusion could help suppress pseudo-label noise.
\end{itemize}
Future research could address these limitations by developing standardized indoor benchmarks, improving pseudo-label generation through uncertainty modeling, and incorporating temporal cues or domain-adaptive learning strategies to enhance robustness and generalization.

\section{Conclusions}
\label{sec:conclusions}
Despite being originally developed for autonomous driving scenes, the ScaLR 2D-to-3D distillation framework shows adaptability to indoor settings, with some limitations. The experiments on four indoor datasets provide initial evidence that 2D-to-3D distillation can effectively learn 3D feature representations, without manual annotations. These findings highlight the potential of cross-modal distillation for indoor lidar scan segmentation feasibility studies.

The cross-dataset results reveal domain gaps caused by sensor variations, as different lidars have different scan patterns, and scene variations, as different buildings and environments appear different. However, the framework remains robust with respect to changing the VFM model, which is likely because the lidar scan representation of the scene and the related embeddings are saturated so that providing more descriptive richness from the VFM features does not improve the quality of the lidar features.
Overall, this study provides a practical reference for adapting large-scale cross-modal distillation to indoor scenes, and its performance.

\section*{Acknowledgements}
This work was supported by the European Union under the Horizon Europe RIA project XTREME (Grant No. 101136006).

{
	\begin{spacing}{1.17}
		\normalsize
		\bibliography{ISPRSguidelines_authors} 
	\end{spacing}
}

\end{document}